\documentclass[journal]{journal}
\ifCLASSINFOpdf
\else
\fi
\usepackage{graphicx}
\usepackage{caption}
\usepackage{subcaption}
\usepackage{svg}
\usepackage{amsmath}

\begin{document}

\title{\LARGE \bf
Controller Design and Experimental Evaluation of a Motorized Assistance for a Patient Transfer Floor Lift
}

\author{Donatien Callon, Ian Lalonde, Mathieu Nadeau and Alexandre Girard
\thanks{D. Callon, I. Lalonde and A. Girard are with the Faculty of Engineering, Université de Sherbrooke, Sherbrooke, QC J1K 0A5, Canada (alexandre.girard2@usherbrooke.ca)}
\thanks{M. Nadeau is with Orixha, Liquid Ventilation for Life, 94700, Maisons-Alfort, France}

}


\maketitle
\thispagestyle{empty}
\pagestyle{empty}

\begin{abstract}
Patient transfer is a challenging, critical task because it exposes caregivers to injury risks. Available transfer devices, like floor lifts, lead to improvements but are far from perfect. They do not eliminate the caregivers’ risk of musculoskeletal disorders, and they can be burdensome to use due to their poor maneuverability. This paper presents a new motorized floor lift with a single central motorized wheel connected to an instrumented handle. Admittance controllers are designed to 1) improve the device maneuverability, 2) reduce the required caregiver effort, and 3) ensure the security and comfort of patients. Two controller designs, one with a linear admittance law and a non-linear admittance law with variable damping, were developed and implemented on a prototype. Tests were performed on seven participants to evaluate the performance of the assistance system and the controllers. The experimental results show that 1) the motorized assistance with the variable damping controller improves maneuverability by 28\%, 2) reduces the amount of effort required to push the lift by 66\% and 3) provides the same level of patient comfort compared to a standard unassisted floor lift.
\end{abstract}


\section{INTRODUCTION}
\begin{figure}[thpb]
     \centering
     \includegraphics[scale=0.38]{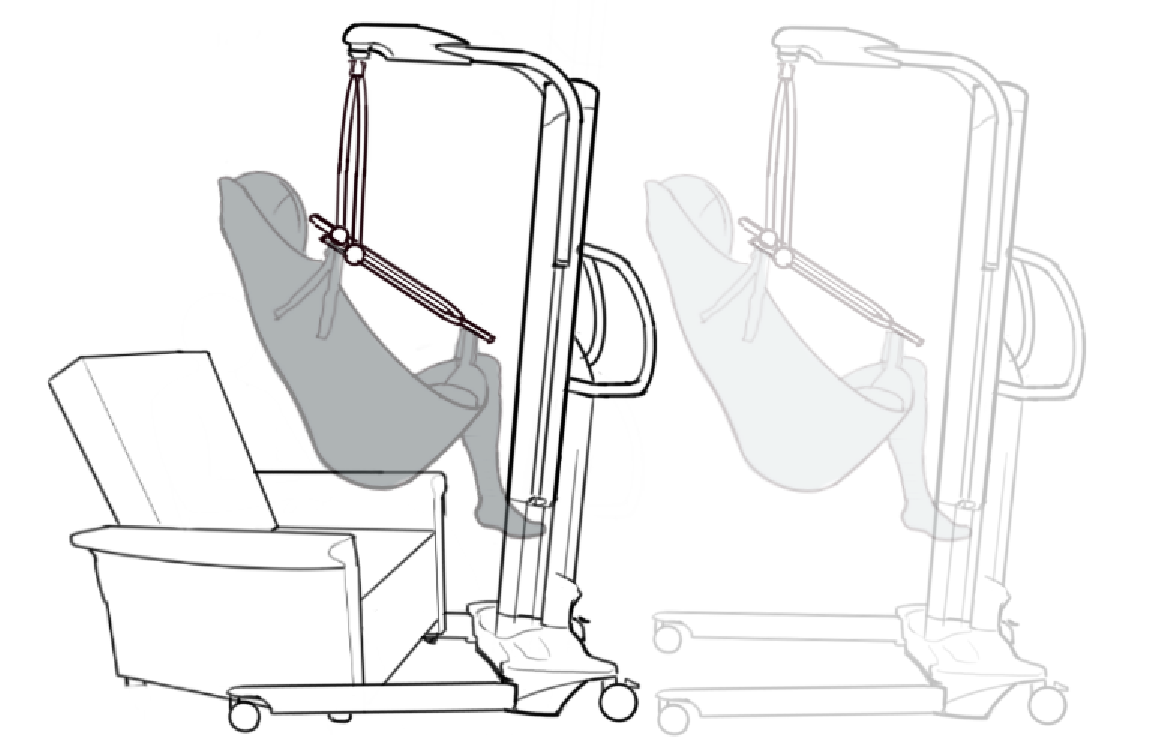}
     \caption{Patient transfer with a floor lift}
     \label{fig:MaxiMove}
     \vspace{-15pt}
\end{figure}

\IEEEPARstart{T}{he} aging population and the worldwide increase in body weight have created new issues in the medical and paramedical sectors. The caregiver profession is associated with many musculoskeletal disorders due to the considerable amount of effort required in a workday \cite{davis_prevalence_2015}. The transfer of a passive patient, one who is unable to move independently from one place to another, is one of the most redundant and difficult tasks for caregivers and is partly responsible for their back disorders \cite{callison_identification_2012}. The National Institute for Occupational Safety and Health (NIOSH) conducted different studies to better understand the development of back disorders. The safety limit for the spinal column is 3,400 N in compression and 1,000 N in shear \cite{daynard_biomechanical_2001}; they are often exceeded during patient-handling tasks \cite{marras_lumbar_2009}. To relieve the caregivers, hospitals and retirement houses are equipped with assistance devices, such as the floor lifts (Figure. \ref{fig:MaxiMove}).
Although these devices have many advantages over unassisted methods for transferring patients, they do not necessarily protect the spinal column. Some medical experiments were performed to measure the effort some users put on their spinal column during a transfer using a floor lift, and it has been observed that the NIOSH’s limits were still exceeded \cite{knapik_spine_2009}. 

To reduce the amount of effort required for a transfer, motorized floor lifts are now proposed by the industry, for example Joy in Care’s TouchMove \cite{noauthor_powermove_nodate} or Handicare’s EvaDrive \cite{noauthor_evadrive_nodate}. The integration of motorized systems in other medical devices, such as beds or wheelchairs, has already been proven successful \cite{guo_design_2018}, because it has decreased the amount of effort required to push these devices, thus improving their ease of use. However, it is still a challenge to integrate intuitive user-interfaces and adapted control schemes so the assisted devices are adopted and useful for the end-users.

This paper discusses the design and experimental evaluation of an admittance controller for the specific application of the fifth-wheel motorized floor lift (see Fig. \ref{fig:Shéma-Lift}) in a realistic scenario of maneuvering a patient in a small bathroom. Three main performance criterion are considered: 1) the maneuverability, 2) the efforts required to move the floor lift, and 3) the comfort and safety of the patient. Admittance controllers are widely used in the context of human-machine interfaces including assistance devices. For instance, Solea et al. proposed a control strategy centered on the comfort of the person in a motorized wheelchair \cite{solea_robotic_2008}. Duchaine et al. recommended a strategy in which the user’s maneuverability and control are improved through a variable controller \cite{duchaine_stable_2012}. Rosen et al. based their control strategy on myosignals to reduce the required effort for a task \cite{rosen_myosignal-based_2001}. Guo et al. presented an admittance control to command a motorized bed, which improves maneuverability and reduces the effort required to move the bed, but no comparison is made with other control strategies \cite{guo_design_2018}. Finally, many articles focus only on the ergonomic aspect of floor lift, but not on the integration of motorized assistance \cite{reid_learning_2007}. Compared to most of the previous related works, in this article the controller design is conducted considering the 3 performance criterion simultaneously and provide an experimental comparison of multiple control approaches. Finally the presented results are original since the fifth-wheel motorized floor lift is a novel concept \cite{jonsson_jorgen_patient_nodate}.

\section{REQUIREMENTS}
The design of the motorized assistance controller is driven by three main performance aspects: maneuverability, effort and patient comfort.

\subsection{Maneuverability}
\label{section:User's maneuverability}
The lift's maneuverability is an important aspect for the system adoption. Furthermore, a poorly maneuverable lift will probably lead indirectly to increased effort and decreased comfort due to the user having to conduct many correcting motion to achieve its goal. There is no universal metric to measure the maneuverability. In a similar context, Duchaine et al. counted the number of overshoots when a user makes a predefined path to compare the maneuverability from one controller to another \cite{duchaine_stable_2012}. In the same way, Guo et al. counted the number of collisions and the time taken to perform a transfer \cite{guo_design_2018}.
In this study, inspired by those previous works, the maneuverability will be quantified with two metrics: 1) the number of overshoots and 2) the time required to perform a transfer, on the simulated path (see Fig. \ref{fig:Path}).

\subsection{Effort}
One important performance goal is to minimize the effort required to push the lift. The ISO 10535:2006 standard  \cite{noauthor_iso_nodate-1} imposes a maximum starting force of 160 N and a maximum driving force of 65 N. Moreover, to protect the spinal column, there should be no more than 3,400 N in compression and 1,000 N in shear, on the spinal column, which is correlated with applying torques, for instance to turn the floor lift. In their publication, Weston et al. recommended a limit of 66.3 Nm of hand torque to avoid exceeding the spinal column’s limits when pushing a wheelchair \cite{weston_wheelchair_2017} which is a very similar scenario to pushing a floor lift. 

\subsection{Patient's Comfort}
During a transfer, the control strategy must guarantee the comfort of the passenger, who is usually frail and often suffer from cognitive impairment. The motion of the lift must be smooth enough to keep the passenger in safe conditions. The ISO 2631-1:1997 automotive standard \cite{noauthor_iso_nodate} proposes a method to determine the quality of the motion in terms of comfort (Table. \ref{table_Acceleration}). The overall rms acceleration is defined as follows:
\begin{equation}
    a_w=\sqrt{k_x^2a_{wx}^2+k_y^2a_{wy}^2+k_z^2a_{wz}^2}
    \label{eq:comfort}    
\end{equation}
where $a_{wx}$, $a_{wy}$ and $a_{wz}$ are the rms accelerations along the $x$, $y$ and $z$ axes, respectively, and $k_x$, $k_y$ and $k_z$ are multiplying factors. For a seated person, $k_x=k_y=1.4$ and $k_z=1$. The requirement for the lift will be considered to have the rms acceleration under $0.315~m/s^2$, i.e. considered not uncomfortable according to the standard. 

\begin{table}[thpb]
\caption{ISO 2631-1:1997 Standards \cite{noauthor_iso_nodate}}
\label{table_Acceleration}
\begin{center}
\begin{tabular}{|c||c|}
\hline
Overall action & Consequence\\
\hline
$a_w \leq 0.315~m/s^2$ & Not uncomfortable\\
$0.315< a_w \leq 0.63 ~m/s^2$ & A little uncomfortable\\
$0.5 < a_w \leq 0.8 ~m/s^2$ & Fairly uncomfortable\\
$0.8 < a_w \leq 1.25 ~m/s^2$ & Uncomfortable\\
$1.25 < a_w \leq 2.5 ~m/s^2$ & Very uncomfortable\\
$a_w \geq 2.5 ~m/s^2$ & Extremely uncomfortable\\
\hline
\end{tabular}
\end{center}
\end{table}

\section{SYSTEM AND MODELISATION}

\subsection{The Fifth Wheel Concept}
\begin{figure}[thpb]
    \centering
    \includegraphics[scale=0.7]{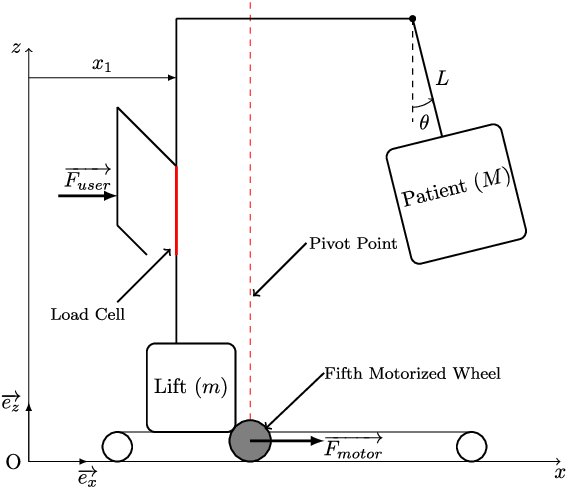}
    \caption{Schematic representation of the motorized fifth wheel concept, composed of a mass m in translation and a suspended mass M in rotation. The user and the motorized fifth wheel are pushing on the lift. A pivot point is created by the fifth wheel, allowing the user to turn the lift easily.}
    \label{fig:Shéma-Lift}
\end{figure}

This investigate controller design specially for the fifth-wheel motorized floor-lift concept. The concept consists of adding a fifth motorized wheel, at the center of the wheel base, to a basic floor lift with four swiveling wheels (see Fig. \ref{fig:Shéma-Lift}). This motorized wheels provide power to assist users while they are pushing forward or backward the lift. The motorized wheel is linked to a load cell that measure the user’s effort on the lift handle, that will be interpreted as an input by the controller. The fifth wheel also has a passive function: it offers a judiciously placed pivot point that allows the user to turn the lift with more ease during patient transfer. Even if the wheel cannot provide torque to rotate the lift, the user will benefit from a lever-arm effect due to the pivot point created by the fifth wheel, as demonstrated in section \ref{section:effort reduction}, by the experimental results.     

\subsection{Dynamic Model}
\label{sec:eom}
To study the behavior of this concept, a simple longitudinal dynamic model is used. It is composed of a first mass (m) representing the floor lift moving in a linear motion and a second suspended mass (M) representing the patient's in the sling. Two horizontal forces act on this system, $F_{motor}$ the propulsion force due to the motorized wheel torque, and $F_{user}$ the force applied on the handle by the user. Parameters of the system are listed at Table \ref{table:Requirements}. Floor lifts are designed to transfer people up to 272 kg. Thus, a high variability of mass $M$ need to be taken into consideration to design the controller.

\begin{table}[thpb]
\caption{System's parameters}
\label{table:Requirements}
\begin{center}
\begin{tabular}{|c||c|}
\hline
Requirement & Value\\
\hline
Maximal speed & 0.8 $m/s$\\
Patient weight (M) & [0-272] kg  \\
Lift weight (m) & 100 kg\\
Length of the pendulum (L) & 0.5 m\\
\hline
\end{tabular}
\end{center}
\end{table}

The equation of motion on this simplified longitudinal model are given by:
\begin{equation}
\begin{cases}
ML^2\Ddot{\theta} + ML\Ddot{x}_1cos(\theta) + MgLsin(\theta) =0 \\
(m+M)\Ddot{x}_1+ML\Ddot{\theta}cos(\theta)-ML\dot\theta^2sin(\theta)  = F_{ext}\\
F_{ext} = F_{user} + F_{motor} + F_{perturbation}
\end{cases}
\end{equation}

\section{Controller Design}
\subsection{Control scheme selection}

\begin{figure}[ht]
     \centering
     \includegraphics[scale=0.42]{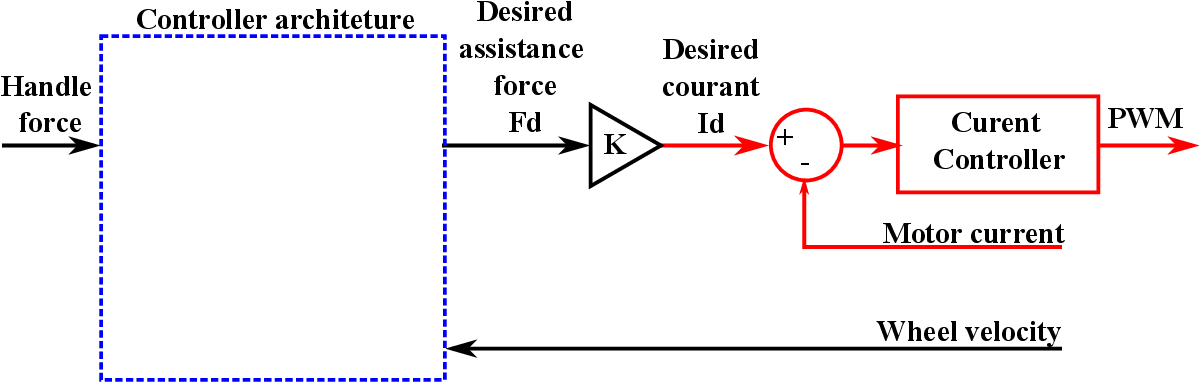}
     \caption{Architecture}
     \label{fig:control}
 \end{figure}

The controller needs to specify the desired motor force $F_d$, as a function of the force read at the handle $F_{user}$ and the wheel velocity $\dot{x}_1$ that is also measured and available to the controller, see Fig. \ref{fig:control}. The desired assistance force is enforced at low-level by a high-bandwidth current controller. Note that the angle of the patient balancing in the sling in not measured.

\begin{figure}[ht]
\begin{subfigure}{0.15\textwidth}
  \includegraphics[width=\linewidth]{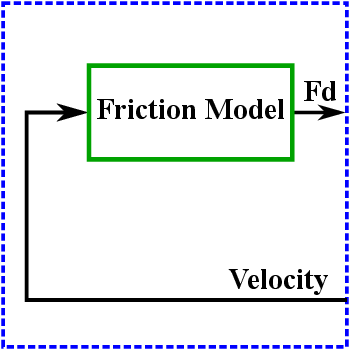}
  \caption{Friction compensation}
  \label{fig:Friction_Compensation}
\end{subfigure}\hfil 
\begin{subfigure}{0.15\textwidth}
  \includegraphics[width=\linewidth]{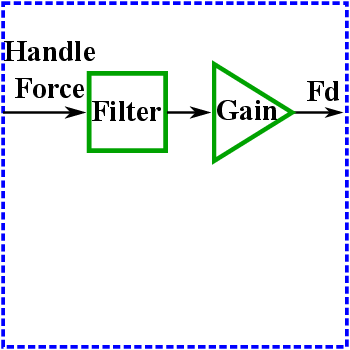}
  \caption{Force amplification}
  \label{fig:Torque_control}
\end{subfigure}\hfil
\begin{subfigure}{0.15\textwidth}
  \includegraphics[width=\linewidth]{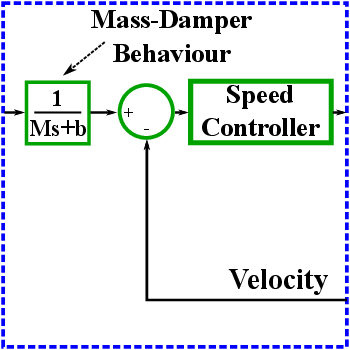}
  \caption{Admittance control}
  \label{fig:speed_control}
\end{subfigure}\hfil 
\caption{Three options of control approaches}
\label{fig:architecture_comparison}
\end{figure}

 Three main families of approaches were developed and compared in a preliminary phase, see figure \ref{fig:architecture_comparison}.
\textbf{1)} The first option (Figure \ref{fig:Friction_Compensation}), a friction compensation controller, estimates the resistive force at the wheel using a velocity measurement and compensates it with the motorized wheel \cite{sehoon_oh_sensor_2005}. This approach has the advantage of allowing the user to push on the system at any contact point, since the user's force measurement is not required. However, the approach has drawbacks and risk issues, to which the floor-lift application is particularly sensitive. The variety of external conditions, such as the floor type (carpet, concrete, parquet) or the weight of the patient, makes it difficult to precisely estimate the resistive force at the wheel. This could lead to dangerous undesired behaviors from the lift in unforeseen situations. \textbf{2)} The second option (Figure \ref{fig:Torque_control}), is a force amplification scheme. Leveraging the instrumented handle, the user's force is measured and filtered in real time. The motor is then asked to provide an assistance force ($F_d$) proportional to the user's force, i.e. thus amplifying the provided user force at the handle. \textbf{3)} The third option (Figure \ref{fig:speed_control}), is an admittance control scheme that uses both sensors (wheel velocity and user's force). A velocity controller is used at low levels to impose a wheel motion at high bandwidth. The desired wheel velocity is computed by an outer loop that consists of a virtual dynamic model (mass-damper) for which the input is the user force at the handle.


After preliminary implementation and testing, option 1 was set aside for this particular application since the risk of instability would be dangerous in the context of transporting patients. Options 2 and 3 are safer in the sense that the motor only react to a direct user input. With option 2, the force amplification scheme directly converts the force of the user into torque on the wheel. Since the wheel only provide an additional force the lift remains backdrivable if a user tries to move it without using the handle. With option 3, the admittance control, the user’s force is converted into speed through software and the wheel motion is imposed by the controller. This can be less intuitive to user since the link between the user and the motion is more indirect.
However, with this approach, the behavior is much less sensitive to a change in the weight of the patient or a change in friction due to different floor conditions (linoleum, concrete, tile, carpet, etc.). For a given input force of the user, the same final speed will be reached regardless of the external conditions. With the force amplification scheme, the final speed will depend on the friction conditions and the weight of the patient. Simulation and preliminary user tests with a prototype where option 1, 2 and 3 were available confirmed major difference in behavior on various surface for option 1 and 2. The admittance control scheme was thus selected for the study.


\subsection{Admittance Controller Design}
The admittance control law convert the measured user input at the handle $F_{user}$ into a desired lift longitudinal velocity $v_d$ (imposed by a low-level velocity controller). The mapping from force to velocity, that can be seen as a virtual dynamic, is here defined as a simple mass-damper system, defined by the following equation:
\begin{equation}
    F_{user} =M \dot{v_d} + b_0 v_d 
\end{equation}
The equation can also be parameterized in term of a time constant $\tau=\frac{M}{b_0}$ and a gain $K=\frac{1}{b_0}$, and be expressed in the form of a transfer function:
\begin{equation}
    \frac{V_d(s)}{F_{user}(s)} = \frac{K}{1+\tau s}
\end{equation}


The gain $K$ governs the force that the user must apply at a certain velocity, and the time constant $\tau$ controls the transient response during operations, such as change of direction, stopping or starting. The gain $K$
was determined by testing different values with users. It was initially set at a low value, and incremented until the users feel that the motorized assistance was too sensitive. 



The time constant was chosen to be as sensitive as possible while respecting the patient comfort criteria given by the maximal acceleration of 0.315 m/s$^2$ (the recommended limit of ISO-2631-1:1997 standards). Numerical simulations, based on the presented equation of motion in section \ref{sec:eom}, were used compute the acceleration response of the patient with respect to worst case sudden step user inputs. It was found that the swinging behavior of the patient in the sling is the limiting factor, i.e. the reactivity of the lift must be limited to avoid too sudden acceleration leading to too large uncomfortable balancing motion for the patient.


\subsection{Diminution of the Patient's Swings: An Unsuccessful Approach}

In order to improve the patient comfort by decrease the balancing motion, two advanced approaches were evaluated. The first approach was to study the integration of a partial feedback linearization (PFL) control strategy \cite{spong_partial_1994}. Instead of imposing the velocity of the lift, with the PFL approach it is possible (leveraging the knowledge of the equation of motion) to instead impose the horizontal velocity of the patient. Simulations assuming full state feedback showed that it is theoretically possible to totally attenuated patient balancing motions. However, this strategy is hard to implement in practice. It would require estimating the un-measured angular position of the pendulum representing the patient and many parameters in the equation of motion.

\begin{figure}[!htb]
    \centering
    \begin{subfigure}{0.45\textwidth}
    \centering
    \includegraphics[scale=0.5]{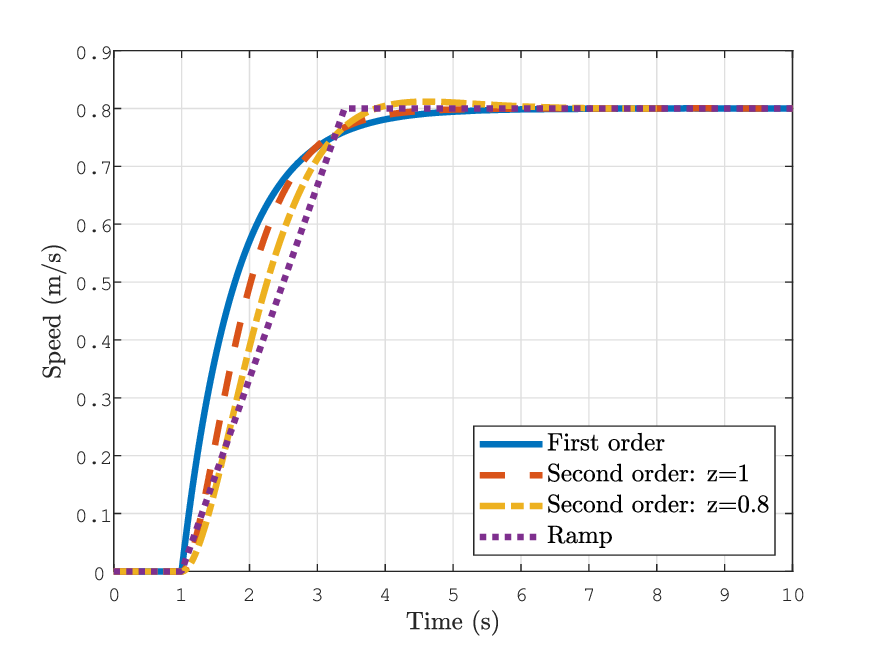}
     \end{subfigure}
     ~
     \begin{subfigure}{0.45\textwidth}
     \centering
    \includegraphics[scale=0.5]{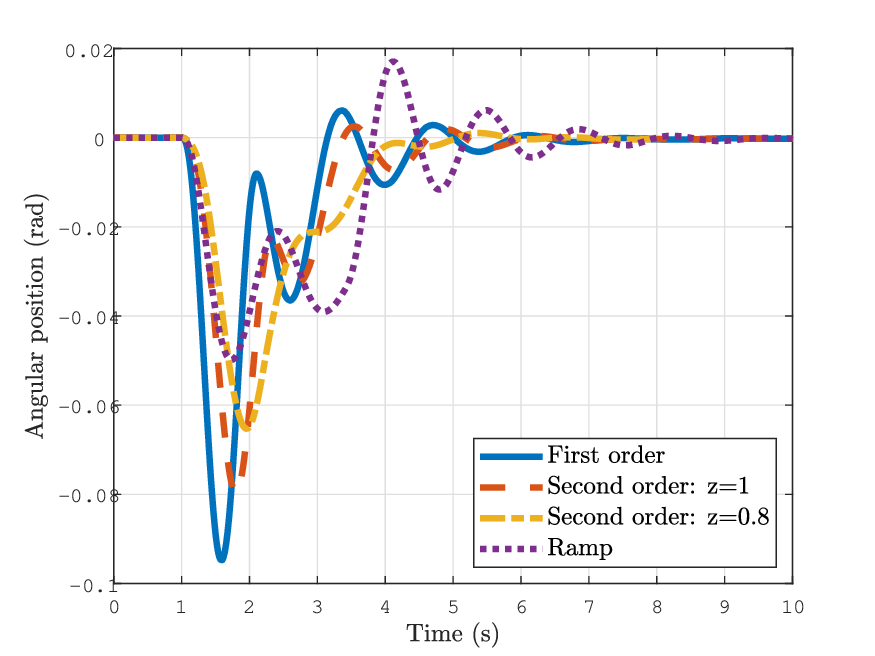}
    \end{subfigure}
    \caption{Comparison of the different angular responses depending on which speed command is sent to the motorized wheel}
    \label{fig:angular_response}
\end{figure}

A simpler balancing-minimizing approach not requiring as many estimated parameter was also investigated. The idea is to  emulate the PFL controller behavior based on time filtering instead of state-feedback. Second-order admittance law were found to minimize the jerk when the lift is starting to lower the excitation of the pendulum balancing mode. Figure \ref{fig:angular_response} shows the different desired speed $v_d$ command and the angular responses $\theta$ of the patient, in response to a step input of user force $F_{user}$.


It appeared that with a second-order admittance law, it was possible to reduce the balancing motion and thus decrease the overall acceleration felt by the patient. However, some preliminary tests, involving three participants trying to follow a path with a second-order admittance controller, revealed that the second-order behavior is not as intuitive as the first order mass-damper behavior. Furthermore, during the tests some users lost control of the lift. The main hypothesis for this results is that the increased delay between a user input and the lift reaction lead to the user trying to overcompensate, which lead an unstable system when including the human user in the loop. It was concluded that it is best to leave the task of filtering the user input to avoid large balancing motion in in hand of the human user himself. 


\subsection{Variable Admittance Control}

One limitation of the linear first-order admittance law is that it leads to a big trade-off between tuning for the required effort when cruising at a constant velocity and the transient response. A variable controller, consisting of changing the admittance law coefficients depending on the situation, was thus investigated to adapt the damping coefficient to the user’s intentions \cite{duchaine_stable_2012}.

When the lift is going in one direction and the user is pushing towards the other, he intends to change the direction of the lift. The lift needs to promptly decelerate, which means that the controller needs a high-damping ratio. 
\begin{equation}
    V\times F_{user} < 0 \Rightarrow B \big\uparrow
\end{equation}
Also, when the lift is going towards a specific direction and the user is pushing in the same way, he intends to go faster. The lift must accelerate, and the controller should use a small damping coefficient. 
\begin{equation}
    V\times F_{user} > 0 \Rightarrow B \big\downarrow
\end{equation}
Thus, it is proposed to modify the damping coefficient through a linear equation depending on the sign of the speed and the amplitude of force.
\begin{equation}
b=\alpha+\beta \times sign(V)\times F_{user}
\end{equation}
The $\alpha$ and $\beta$ coefficients were tuned to get a good behavior of the lift at low speeds and to have the same sensitivity at maximal speeds than the linear admittance controller (30~N to get to 0.8~m/s). However, the damping coefficient should be limited to prevent the lift from being too sensitive and hard to control \cite{duchaine_stable_2012}. 
The coefficients of the varying controller were parameterized using $b_0$, the damping coefficient of the classical admittance controller, to compare the classical admittance controller and the variable controller. 
$\alpha$ was set to $6b_0$ to increase the lift's reaction at low speed and when changing directions. $\beta$ was established to get 0.8~m/s when the user is pushing at $F_0=30~N$, and thus obtain the same speed range than the baseline admittance controller.

\begin{equation}
\begin{array}{cc}
    \frac{V(s)}{F_{user}(s)}=\frac{1}{Ms+b}  \\
    \\
    \begin{cases}
    b=6b_0-sign(V)\frac{5b_0}{F_0}F_{user} ~if~|F_{user}|<F_0\\
    b=b_0~if~|F_{user}|>F_0
    \end{cases} 
\end{array}
\end{equation}

\begin{figure}[thpb]
    \centering
    \includegraphics[scale=0.52]{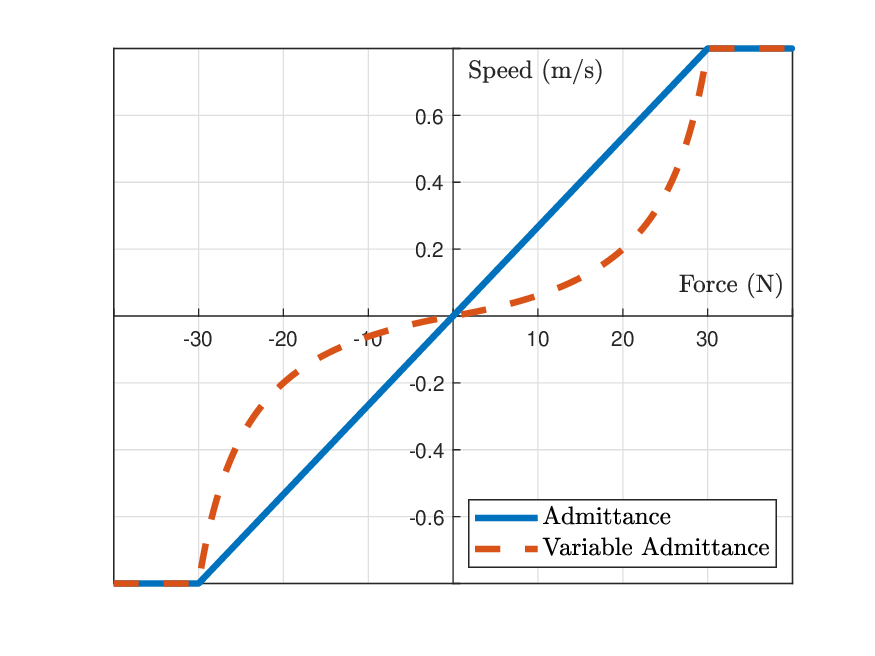}
    \caption{Final velocity versus the input force for an admittance controller and a variable admittance controller}
    \label{fig:adaptive_damping}
\end{figure}

Figure \ref{fig:adaptive_damping} shows the final velocity versus the input force for the baseline admittance controller and the variable damping controller, following equations \ref{eq:ClassicalVsVariable}. The non-linearity generated by the varying damping coefficient creates a better resolution of the user-interface at low speed: the controller requires larger increments of force to adjust the speed. This should helps to make small adjustments of the lift, and it is potentially the factor that explain the measured improvement to the maneuverability in the experimental evaluation.




\begin{equation}
\label{eq:ClassicalVsVariable}
    \begin{cases}
    V_{admittance}=\frac{F_{user}}{b_0}\\
    V_{variable~admittance}=\frac{F_{user}}{6b_0-sign(V)\frac{5b_0}{F_0}F}
    \end{cases}
\end{equation}

\section{PERFORMANCE EVALUATION}

\subsection{Testing Protocol}
\begin{figure}[thpb]
    \centering
    \includegraphics[scale=0.32]{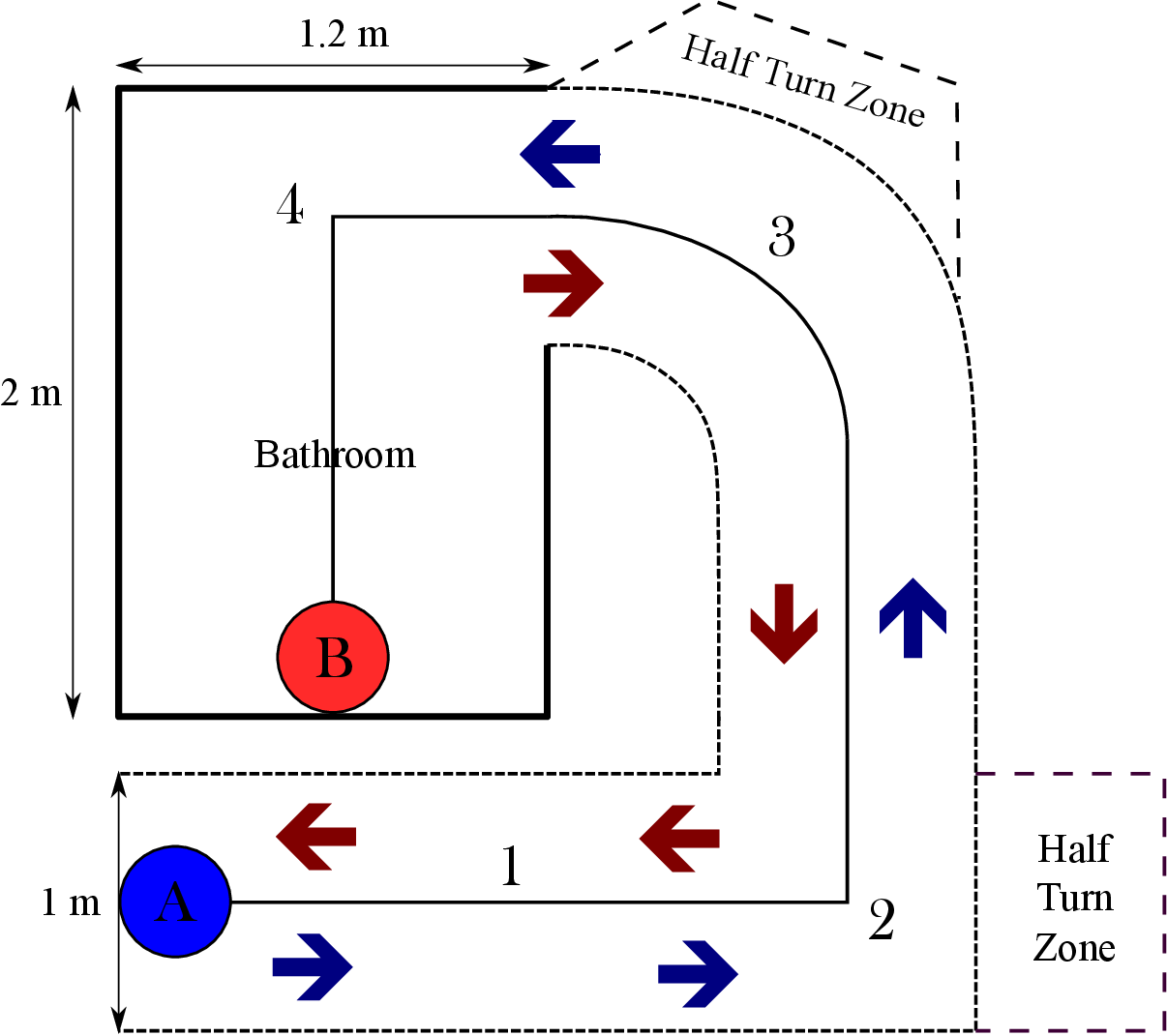}
    \caption{Patient handling path. The path is composed of 1) a straight section, 2) a sharp turn section, 3) a gradual turn section and 4) a confined turn. The dotted lines represent the tape on the floor that was used to count the number of overshoots. The solid black lines represent the wall of the simulated bathroom.}
    \label{fig:Path}
\end{figure}

The performance evaluation was conducted with seven participants: four men and three women aged 31–56 years old, with the average at 40 years old. Each participant was asked to execute a series of 12 patient transfer simulations with different simulated patient weights and different controllers. The transfer consists of moving a simulated patient from a chair to a simulated bathroom and back to the chair. The goal is to simulate a real case of a patient transfer with the same restrictions in terms of space, positioning and path. The path is inspired from Marras’ article \cite{marras_lumbar_2009}. It is made of a straight part, a sharp turn, a long turn and a second sharp turn when the user is entering a bathroom (Figure \ref{fig:Path}). The user starts at zone A as if they were picking up someone from a chair. The user starts in a backward position and must back up to the half-turn zone and then turn and walk forward to zone B. Once at B, they can go back to A by using the second half-turn zone again, walking backward to the half-turn zone and then forward to A. Three wooden dummies with a humanoid shape were filled with cast-iron weights and used as simulated patients. They weighed 80~kg, 130~kg and 180~kg, respectively.

The participants tested four different cases. 
The first one was with no assistance at all and no fifth-wheel present, as if it was a basic floor lift. 
The second one was with the help of the fifth wheel but without any active assistance. The motorized wheel was used but unplugged from any electrical source, and it served as a basic wheel. The goal was to see the passive effect of the fifth wheel. The third case was with the motorized wheel controlled by the baseline admittance law. 
The last case was was with the motorized wheel controlled by the varying admittance law. The order in which the weights and cases were tested was established randomly. The participants were not aware of the simulated patient weight to avoid any cognitive bias on their perception of the lift. The running order of the controllers was also determined randomly for each participant. They were not told which controller mode they were testing to avoid any cognitive bias.

\subsection{Measurement of the performance}

\subsubsection{Maneuverability}
Maneuverability was measured with two metrics: 1) the time needed to accomplish a run and 2) the number of overshoots. Some tape was put on the floor to evaluate the amount of space needed to execute the path (dotted lines on Figure \ref{fig:Path}). Each time the user or the lift stepped outside the zone delimited by the tape on the floor or made a collision with the walls, it was counted as one overshoot. The participants did not know that the number of steps outside the zone was counted to make sure they do not force themselves to stay in the zone and avoid any cognitive bias. 

\subsubsection{Effort}
The effort reduction was quantified through a three-axis load cell. This load cell was made from a specially designed instrumented tube with 10 strain gauges (8 linear and 2 shear) that could measure the longitudinal force as well as the torque on the lift handle. Each axis was calibrated prior to experiments. The acquisition of the strain bridges was made with the HX711 (Avia, China) load cell amplifiers at a 10~Hz frequency. The acquisition board was an Ardbox (Industrial Shield, Spain), and the recording was made directly on a laptop through a serial USB connection. The integral of the squared force and the integral of the squared torque during each run were used as final metric for the effort:
\begin{equation}
        \int F^2 dt \quad \quad
        \int T^2 dt
    \label{eq:squared force}
\end{equation}

\subsubsection{Patient Comfort}
Patient comfort was evaluated by measuring the acceleration on the dummy with an inertial measurement unit. An MPU-6050 (TDK InvenSense, United States) was used with a frequency of 200 Hz and a moving average of 50 points, and it was linked to an Arduino Mega. The values were recorded on a laptop through an HC-05 (DSDTech, China), Bluetooth to serial module. The data were post-treated to calculate the overall acceleration, using eq. \ref{eq:comfort}, as previously described, to be used as a performance criterion.

\subsubsection{Qualitative Feedback}
In addition to the 3 quantitative metrics, after each run, the participants were asked to review the transfer. They were asked to give grades on a scale of 1–10 about their appreciation of the effort (10 being no effort needed), their feeling of control (10 meaning being in full control) and their global appreciation. The participants could ask for revised grades after the transfers.

\begin{figure*}[htb]
    \centering 
\begin{subfigure}{0.3\textwidth}
  \includegraphics[width=\linewidth]{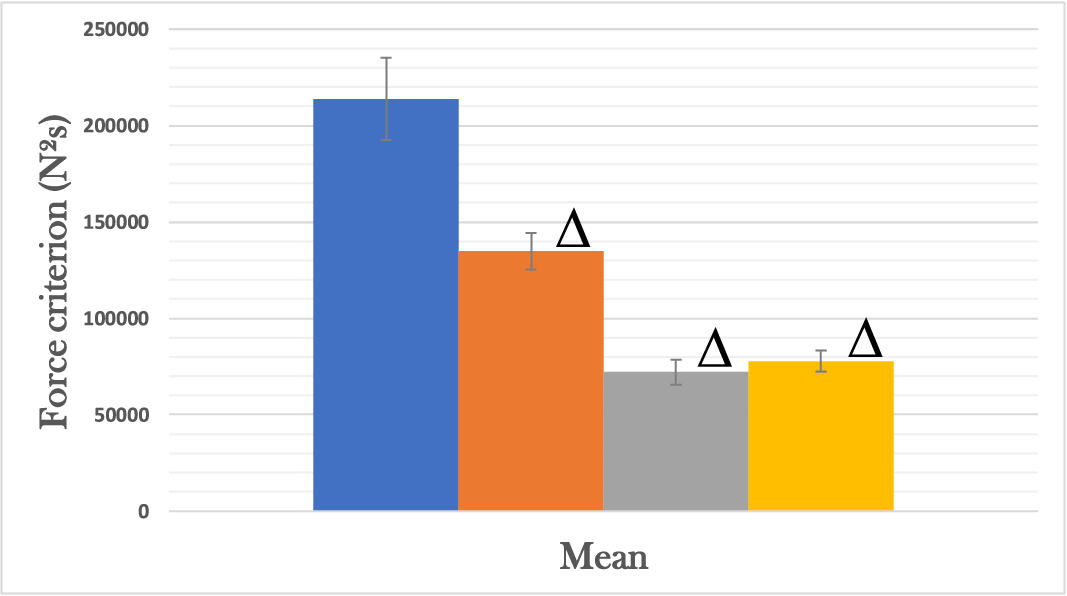}
  \caption{Force criterion $\int F^2 dt$}
  \label{fig:force}
\end{subfigure}\hfil 
\begin{subfigure}{0.3\textwidth}
  \includegraphics[width=\linewidth]{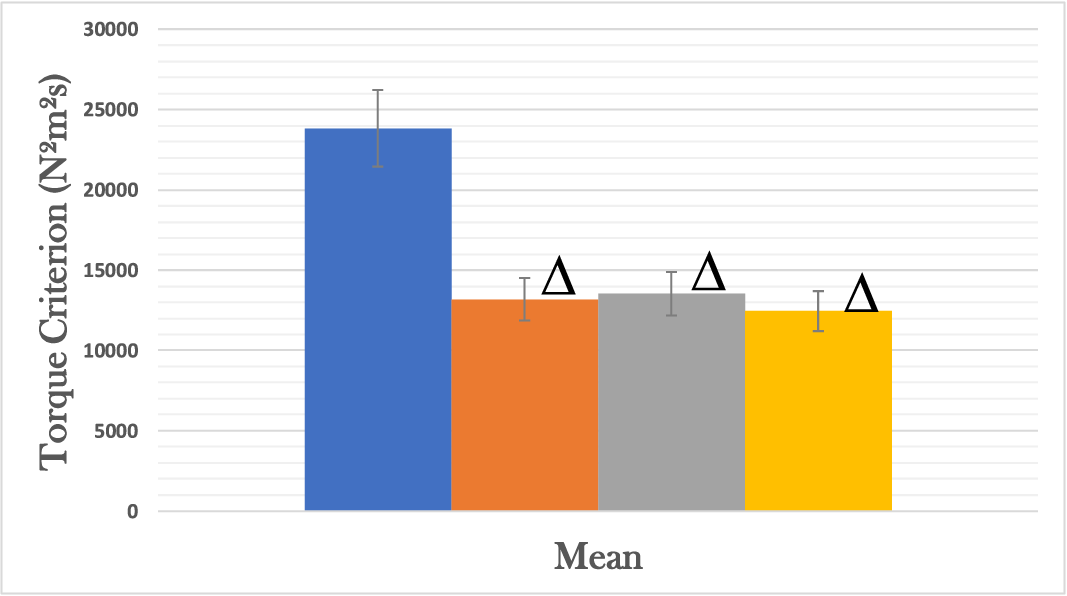}
  \caption{Torque criterion $\int T^2 dt$}
  \label{fig:torque}
\end{subfigure}\hfil 
\begin{subfigure}{0.3\textwidth}
  \includegraphics[width=\linewidth]{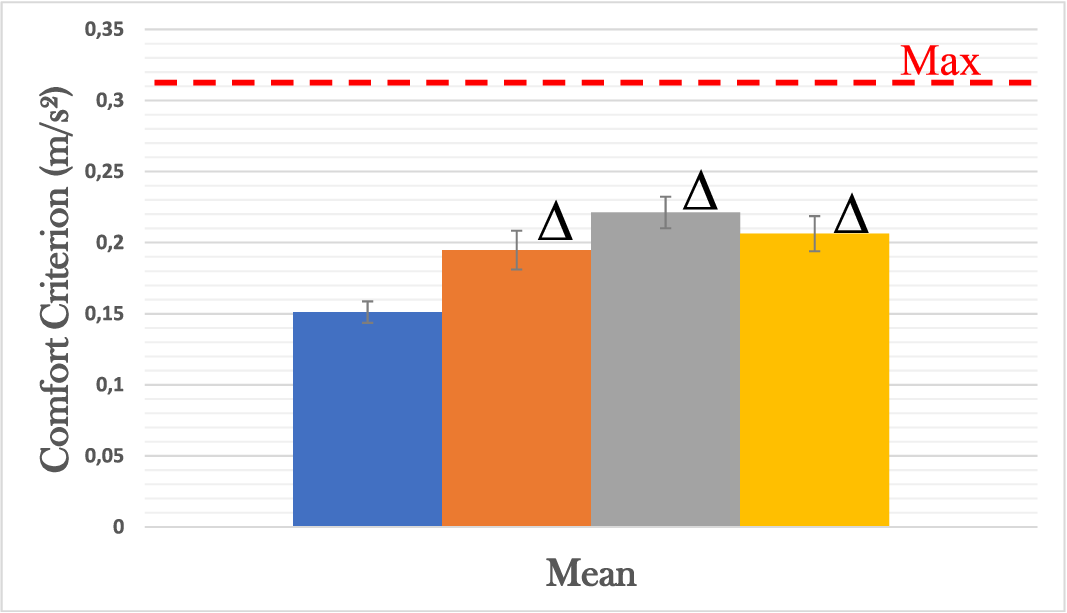}
  \caption{Comfort criterion (Eq. \ref{eq:comfort})}
  \label{fig:Comfort}
\end{subfigure}
~
\medskip
\begin{subfigure}{0.3\textwidth}
  \includegraphics[width=\linewidth]{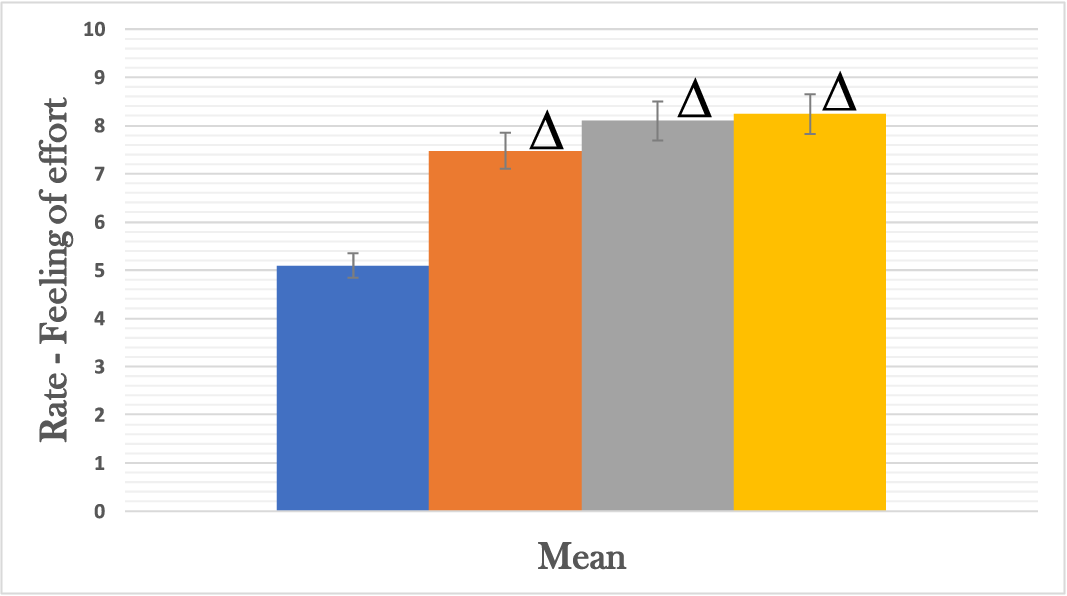}
  \caption{Appreciation of effort}
  \label{fig:feeling_effort}
\end{subfigure}\hfil 
\begin{subfigure}{0.3\textwidth}
  \includegraphics[width=\linewidth]{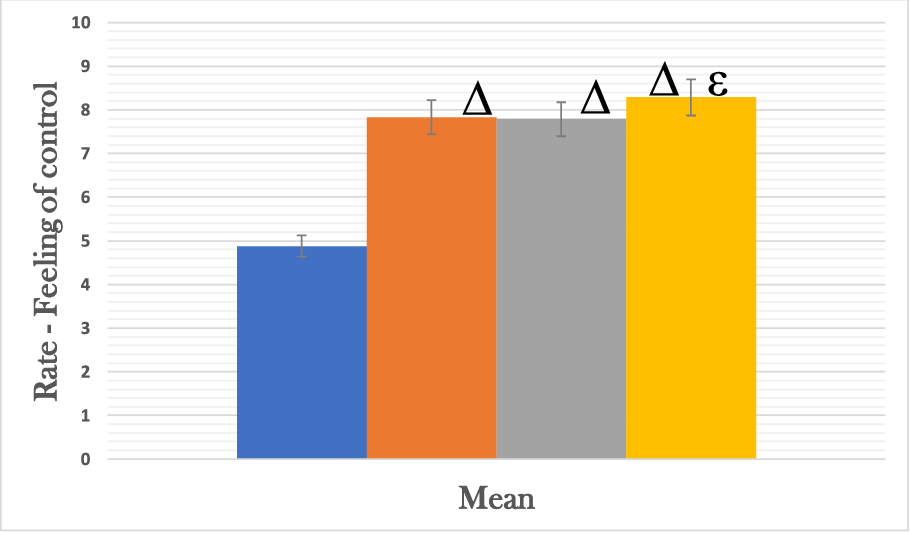}
  \caption{Appreciation of control}
  \label{fig:feeling_control}
\end{subfigure}\hfil 
\begin{subfigure}{0.3\textwidth}
  \includegraphics[width=\linewidth]{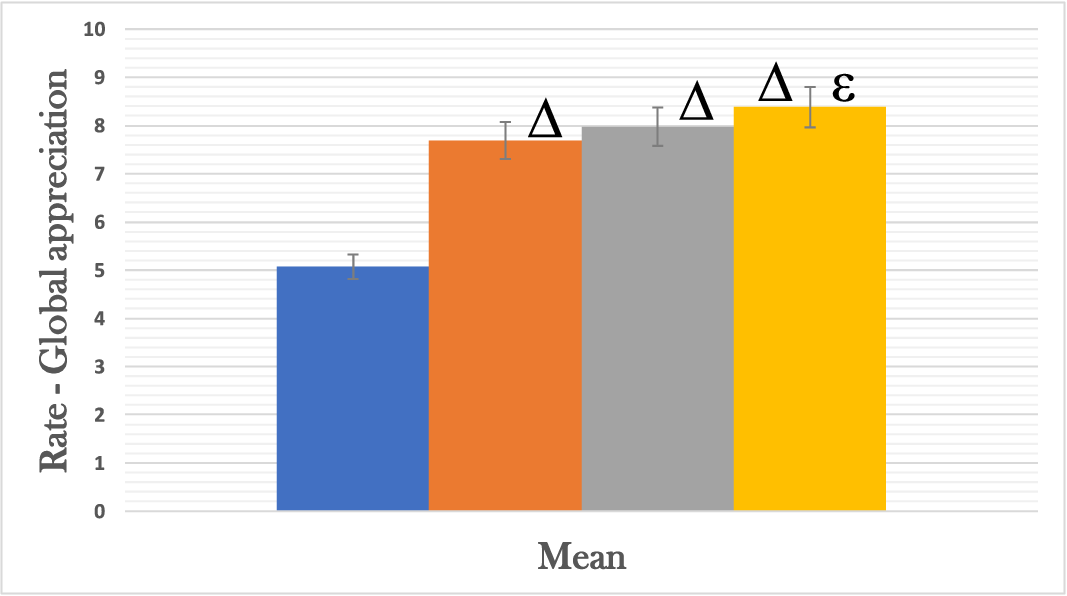}
  \caption{Global appreciation}
  \label{fig:feeling_global}
\end{subfigure}
~
\medskip
\begin{subfigure}{0.3\textwidth}
  \includegraphics[width=\linewidth]{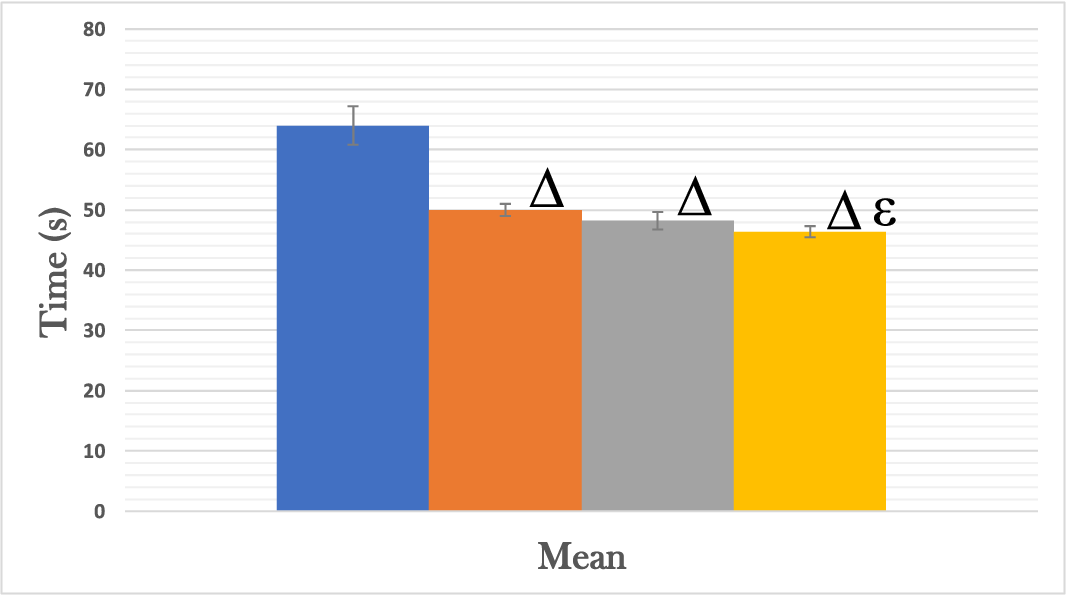}
  \caption{Time to perform the transfer}
  \label{fig:Time}
\end{subfigure}\hfil 
\begin{subfigure}{0.3\textwidth}
  \includegraphics[width=\linewidth]{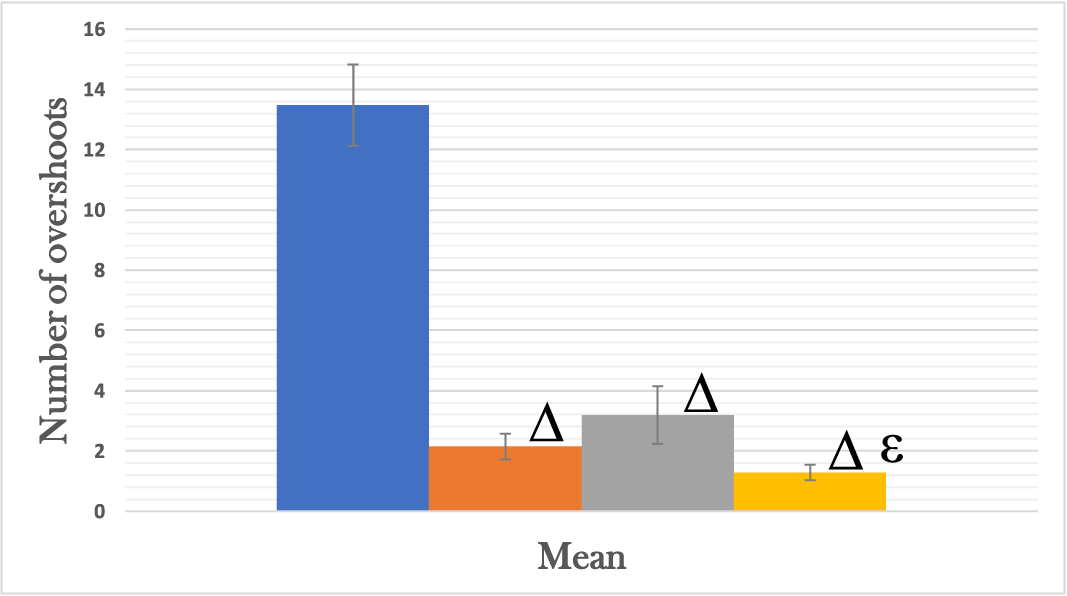}
  \caption{Number of overshoots}
  \label{fig:Overshoot}
\end{subfigure}\hfil 
\begin{subfigure}{0.3\textwidth}
  \includegraphics[width=\linewidth]{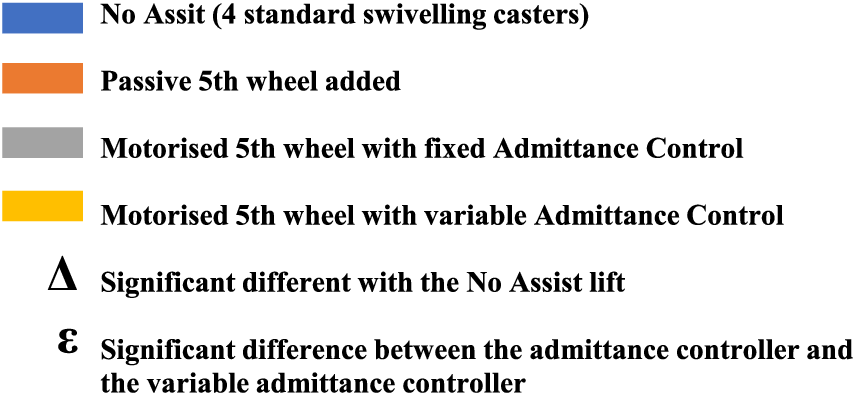}
  \label{fig:legend}
\end{subfigure}

\caption{Comparison of the performances of the different lift configurations}
\label{fig:performances}
\end{figure*}

\section{EXPERIMENTAL RESULTS}

The results are presented in Fig. \ref{fig:performances}. Each column represents the mean, and an error bar indicates the standard deviation of each criterion for all the different runs of the participants. The Student paired t-test was used to determine if the difference in mean values between each lift configuration is significant. The null hypothesis was rejected if  $p-value > 0.1$. Statistical tests were performed using the Excel statistics toolbox (Microsoft, USA).

\subsection{Maneuverability}

Figure \ref{fig:Time} shows the time taken by users to perform the transfers depending on the lift configuration.
The average time to realize the path with the unassisted lift is 64 seconds, whereas it takes 50 seconds with the passive fifth wheel lift.
There is an average improvement of 1.2 seconds between the passive fifth wheel and the baseline admittance controller, and an additional average improvement of 1.8 second with the varying admittance controller. 



Figure \ref{fig:Overshoot} presents the number of overshoots observed during the tests. The unassisted lift has an average number of overshoots of 14 per run, which is more than with a fifth wheel, motorized or not. 
It was observed that the participants tend to change position to push the lift without the fifth-wheel, which leads to an important number of steps outside the delimited zone. 
This behavior becomes laborious when the participant is in a restricted space, for instance, the simulated bathroom.
The variable controller has the lowest number of overshoot (40\% of difference with the passive fifth wheel lift), but the baseline admittance controller was a little worst that using only the fifth-wheel passively.


\subsection{Effort Reduction}
\label{section:effort reduction}

Fig. \ref{fig:force} shows the squared integrals of force for each participant with the different lift configurations. Each bar represents the average of the three runs at 80, 130 and 180 kg. As expected, there was a diminution of effort between the unassisted lift and the motorized lift.
There is an average difference of force of 37\% between the unassisted lift and the passive fifth wheel lift and 66\% between the unassisted and motorized lifts. 
Surprisingly, there is a significant diminution between the unassisted and passive fifth wheel lifts on the force criterion. It was expected that the fifth wheel would only help the user for rotation motion and thus would have only been observed on the torque criterion, which is not the case. 
For the two active cases, the admittance controller and the variable controller have similar levels of force and torque, which is expected since they are tuned to have the same sensitivities (lift speed for a given handle force) at cruising speed.
%
Figure \ref{fig:torque} shows the squared integrals of the torque at the handle for each participant. There is a significant difference of 45\% between the unassisted lift and the other configurations. 
There is no clear difference between using the fifth wheel passively and with the two active controllers. 



\subsection{Passenger Comfort}
Figure \ref{fig:Comfort} is the overall acceleration for each participant. Each bar represents the average of the three different weights. The red dotted line represents a limit of $0.315~m/s^2$ recommended by the ISO-2631-1:1997, which must not be exceeded to ensure a comfortable transfer for the patient. When comparing the acceleration of the different users, it appears that the unassisted lift is more comfortable than the assisted lift. There is a 32\% difference in average between the lift with an admittance controller and the unassisted lift. 
One potential explanation is that when using the lift without fifth wheel, the center of rotation tends to be the patient, while it tends to be the fifth wheel when it is present. 
All controllers have an average acceleration below $0.315~m/s^2$ and can be characterized as comfortable.



\subsection{Qualitative feedback}
Figure \ref{fig:feeling_effort}, \ref{fig:feeling_control},  and \ref{fig:feeling_global} are the different ratings given by the participants on their appreciation of effort and control and their global appreciation of the lifts’ behaviors.

All the users were able to feel a significant difference in terms of effort between the unassisted lift and the passive fifth wheel lift, as well as between the passive fifth wheel lift and the motorized controller. There is 32\% difference between the unassisted and passive fifth wheel lifts, and 8\% between the passive fifth wheel lift and the admittance controller. However, they were not able to note any difference between the baseline admittance controller and the variable admittance controller. 

Regarding the appreciation of control, all participants found that the unassisted lift is challenging to use for transferring a patient in comparison to the others. There is a 38\% rating difference between the unassisted and passive fifth wheel lifts. The admittance controllers and the passive fifth wheel lift were similarly appreciated. 

For the global appreciations, the variable admittance controller was preferred compared to the fixed admittance controller (5\% of difference) for its maneuverability. The admittance controllers are more appreciated than the passive fifth wheel lift (4\% of difference) for its reduction of effort, which matches the comments collected during the activity.


\section{DISCUSSION}
The main purpose of this work was to develop a controller for a motorized fifth wheel added to a patient transfer device and to experimentally validate its performance in a realistic patient transfer simulation. To our knowledge, this is the first study investigating the controller design of a floor lift equipped with a motorized fifth wheel and conducting an extensive experimental validation. The presented motorized fifth wheel concept and the designed variable admittance controller have proven to be significant improvements compared to a regular passive floor lift. 

Thanks to the diminution of the amount of effort required to push the lift, the motorized system should lead to a reduction of injuries for users. The fifth wheel allows a significant reduction of 66\% of force and 45\% of torque at the handle. Without motorized assistance, results show that values between 80 N and 190 N are required to start the lift depending on the weight of the simulated patient, whereas less than 20 N is required with motorized assistance.  Which respects the requirement of a maximum starting force of 160N. Also it is interesting to note that the fifth wheel, even when used passively, leads diminution of the effort, especially in terms of required torque to turn the floor lift, a metric that is correlated with the risk of back injury. The maneuverability is also improved with the motorized assistance. The time required to perform a transfer has been decreased by more than 22\% in comparison to the unassisted lift. Also, the varying admittance control scheme leads to the best maneuverability results in term of all measured metrics.

In this study, it is interesting to note that the conclusions from the measured data are confirmed by the users’ feedback and perception. The difference in effort was measured by the sensors and rated by the participants, as well as the difference in control between each lift configurations. All participants agreed that there is value in the motorization of a fifth wheel for the patient’s transfer lift. 


One limitation of the conducted study is that the participants were not professional caregivers, and some key elements that a real caregiver could have seen might have been missed. Furthermore, it would have been interesting to have a real simulated patient instead of articulated wooden dummies filled with weights. Participants may have behaved differently with a living simulated patient. 


\section{CONCLUSIONS}
This paper present a motorized floor lift concept, discuss the controller design and present an experimental evaluation of multiple performance metrics. Tests with seven participants were performed with different lift configurations: a regular floor lift, a floor lift with a passive fifth wheel,  a floor lift with a motorized fifth controlled by a classical admittance controller and a variable admittance controller The experimental results show that motorized assistance with the variable controller improves maneuverability, reduces the amount of effort required to push the lift by 66\% and preserves the patient’s comfort, in comparison to a standard unassisted floor lift. The variable admittance controller was preferred by the participants in comparison to a baseline linear admittance controller because the system reacts with a more intuitive response to the user’s intentions, which improves the maneuverability.

The next step is to introduce the motorized lift in a hospital environment to investigate the performance when exposed to various real external factors, and to compare the motorized fifth wheel concept to the already existing motorized floor lift solutions. 

\section{Acknowledgements}
This project was financed by CoRoM, a NSERC-CREATE training program specialized in collaborative robotics.
\addtolength{\textheight}{-12cm}   

\bibliographystyle{IEEEtran}
\bibliography{references.bib}
\end{document}